\def\year{2021}\relax
\documentclass[letterpaper]{article} % DO NOT CHANGE THIS

\usepackage{aaai21}  % DO NOT CHANGE THIS
\usepackage{times}  % DO NOT CHANGE THIS
\usepackage{helvet} % DO NOT CHANGE THIS
\usepackage{courier}  % DO NOT CHANGE THIS
\usepackage[hyphens]{url}  % DO NOT CHANGE THIS
\usepackage{graphicx} % DO NOT CHANGE THIS
\usepackage{natbib}  % DO NOT CHANGE THIS AND DO NOT ADD ANY OPTIONS TO IT
\usepackage{caption} % DO NOT CHANGE THIS AND DO NOT ADD ANY OPTIONS TO IT

\usepackage{times}
\usepackage{latexsym}

\usepackage{enumitem}
\usepackage{url}
\usepackage{times}
\usepackage{latexsym}
\usepackage{amsmath}
\usepackage{titlesec}
\usepackage{amssymb}
\usepackage{graphicx}
\usepackage{multirow}
\usepackage{array}
\usepackage{url}
\usepackage{algorithm}
\usepackage[noend]{algpseudocode}

\nocopyright
\author{Anietie Andy,$^1$ Siyi Liu,$^2$ Daphne Ippolito,$^2$ Reno Kriz,$^{2,3}$ Chris Callison-Burch,$^2$ Derry Wijaya$^4$ \\
$^{1}$Penn Medicine, $^{2}$University of Pennsylvania, \\
$^{3}$ Human Language Technology Center of Excellence, Johns Hopkins University, \\ $^{4}$ Boston University \\
{\small {\tt Anietie.Andy@pennmedicine.upenn.edu}, {\tt \{siyiliu,daphnei,ccb\}@seas.upenn.edu}, {\tt rkriz1@jh.edu}, {\tt wijaya@bu.edu}} \\
}

\title{Creating Multimedia Summaries Using Tweets and Videos}

\begin{document}

\maketitle

\begin{abstract}
While popular televised events such as presidential debates or TV shows are airing, people provide commentary on them in real-time. In this paper, we propose a simple yet effective approach to combine social media commentary and videos to create a multimedia summary of  televised events.  Our approach identifies scenes from these events based on %sub-events based on 
spikes of mentions of people involved in the event and automatically selects tweets and frames from the videos that occur during the time period of the spike that talk about and show the people being discussed.
\end{abstract}

\noindent
\begin{center}
\textbf{Introduction}
\end{center}
%\section{Introduction}
\label{introd}

% mention about the motivation being the fact that people like to read live blog posts or listen to live commentators means that we should create a system that can do this automatically from videos and Twitter

\begin{figure}
  \includegraphics[width=0.47\textwidth]{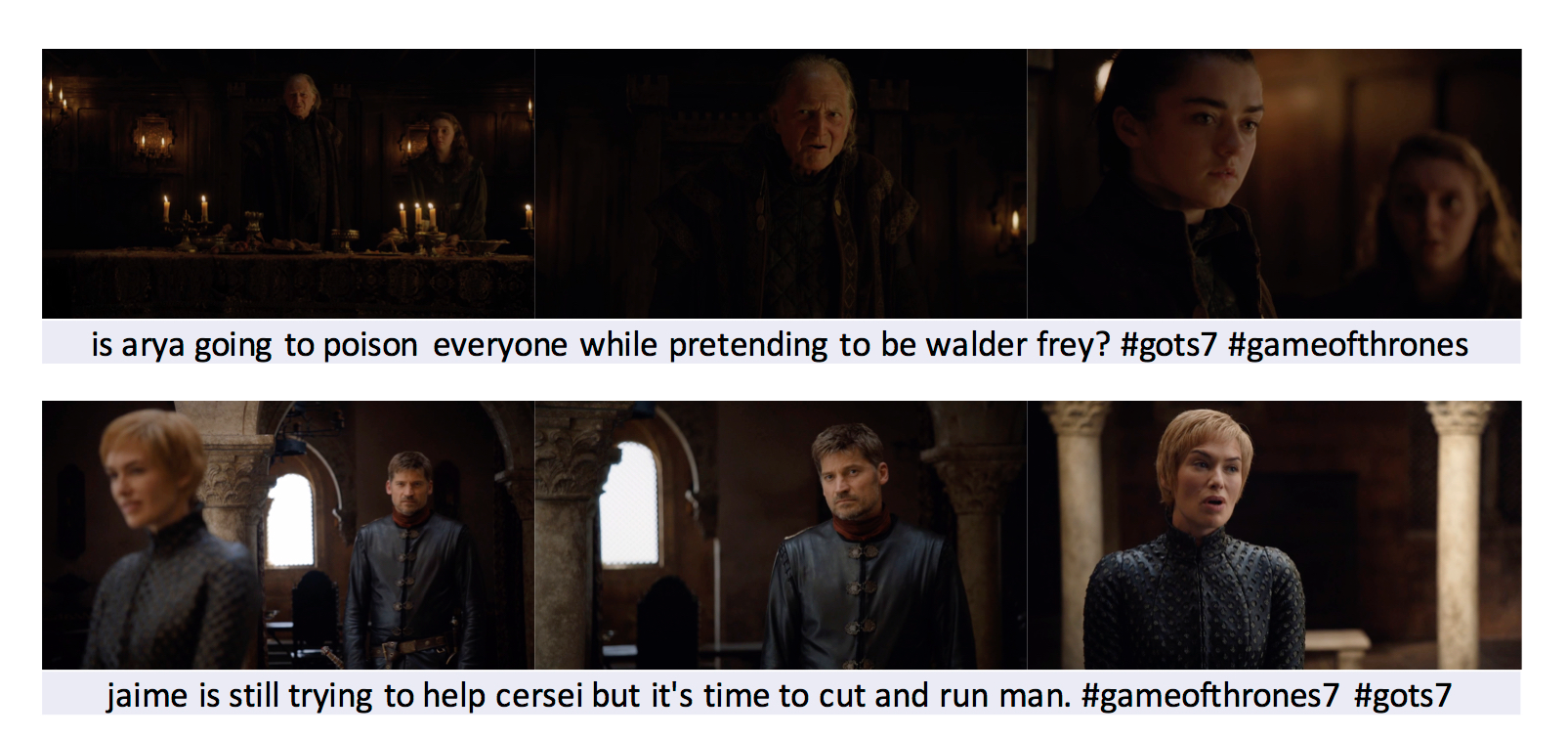}
 \centering
  \caption{Our approach selects tweets paired with key video frames for a multimedia summary of important events in a TV show. Here are tweets about Game of Thrones and their corresponding top-3 video frames selected by our algorithm. 
  \label{small-fig}}
\vspace{-2em}
\end{figure}

Televised events like presidential debates and TV shows %, and sports games 
capture the attention of vast numbers of people--some of whom tweet or discuss about them in real-time. Using user generated contents--UGCs (in social media or discussion forums data such as Twitter or Reddit) to detect important sequential scenes %sub-events 
in a televised event can help with summarizing such events \cite{andy2019winter}, gaining insights about the events \cite{shamma2009tweet}, %resolving pronominal references related to the event \cite{andy2020resolving}, 
or capturing how people react to different parts of an event \cite{shamma2011peaks}. When such events happen, news media often have analysts or commentators discuss various parts of the event; however, automatically generated summaries from UGCs can be useful to capture how individuals are reacting to the event on social media platforms in real time. 
In this paper, we propose a novel, simple yet effective approach that creates multimedia summaries of televised events by selecting social media data (tweets) and video frames of the events that occur in the same timeframe \textit{and} show the people/characters being mentioned in the tweets. Taken together, they provide a multimedia summary (descriptions and pictures) of important parts of a televised event as shown in Figure \ref{small-fig}.%, as shown in Figure \ref{small-fig}.

In televised events, a scene %sub-event -- which can also be referred to as a scene -- 
is defined as "composing of groups of shots that emphasize a specific concept such as a fixed setting" \cite{panda2018nystrom}.
Scenes vary depending on the type of event; for example in a presidential debate, an example of a scene is a candidate responding to a question posed by the moderator, however, in a TV show, a scene could be an interaction between two characters.
%composing of groups of shots, usually emphasizing specific concepts like a fixed setting or the same action \cite{panda2018nystrom}. 
%Sub-events vary depending on the type of event. For example in a TV show, a sub-event could be an interaction between two characters, however, in a presidential debate, an example of a sub-event is a candidate responding to a question posed by the moderator. 
During the broadcast of many televised events, Twitter users create a huge amount of time-stamped and temporally ordered data  \cite{andy2019winter, shamma2009tweet,shamma2011peaks,andy2017constructing,nichols2012summarizing,gillani2017post, andy2020resolving}. We use this tweet-stream data to identify scenes.

%Prior work \cite{shamma2009tweet,andy2019winter} used this tweet-stream data to identify important scenes of an event; similarly, in this work, we use tweet-stream data to identify scenes. 

%We use this tweet-stream data to identify scenes.%sub-events.

%\textbf{TODO: motivate why tweets are used for scene detection instead of video}.\\

Some previous works have developed algorithms to detect scenes from videos related to events \cite{li2020recurrent, vats2020event,einfalt2020decoupling}. In this work, rather than use videos, we are interested in using Twitter data related to events to identify the sequential scenes related to the events. We are interested in this approach because previous works that identified scenes of events from Twitter data related to events focused on identifying the important scenes of these events \cite{shamma2009tweet, andy2019winter}. %Past work used the volume of tweets related to real-time events to identify important scenes of these events \cite{shamma2009tweet, andy2019winter}. 
However, these events also have scenes 
that may not be as important. %Typically, events have few highlights, however, these events have several sub-events. 
In this work, we aim to identify all the scenes 
(important or not) in a given event from Twitter data related to the event. In \cite{andy2017constructing}, it was determined that
in tweets (related to an event) published during the time-period of a scene, %sub-
there is an uptick in the number of mentions of \textit{characters} involved in the scene 
that can be used to summarize the event \cite{huang2018event}. %during a sub-event, there is an uptick in the number mentions of characters involved in the sub-event in Tweets related to the event. 
Taking advantage of this, our approach first identifies scenes
by identifying changes in the minute-by-minute frequency of character mentions in tweets. After identifying a time-range for a scene, 
our approach selects the tweet that best describes the scene 
and the key frame from the video 
that have faces of the characters mentioned in the scene
tweets.

The main contributions of this paper are: %(1) We develop an approach that uses contemporaneous Twitter commentary about a televised event and the people involved to segment the event into sequential scenes (2) We present a novel approach that uses weak supervision data and incremental learning to effectively recognize faces in video frames. Our method outperforms previous weakly supervised approaches on the task of face recognition. %while potentially robust to racial biases. 
%We use this method to select video frames for the identified scenes, thus creating multimedia summaries of the event (3) We collect two new datasets to test our approach.  They include video frames and tweets corresponding to night 1 of the first United States (US) Democratic party presidential debates in 2019 and 7 episodes of the Game of Thrones (GoT). We will release these datasets and code upon publication.

%(1) We develop an approach that uses contemporaneous Twitter commentary about a televised event and the people involved to segment the event into sequential sub-events. (2) We present a novel approach that uses weak supervision data and incremental learning to effectively recognize faces in video frames. Our method outperforms previous weakly supervised approaches on the task of face recognition while potentially robust to racial biases. We use this method to select video frames for the identified sub-events, thus creating multimedia summaries of the event.  (3) We collect two new datasets to test our approach.  They include video frames and tweets corresponding to night 1 of the first United States (US) Democratic party presidential debates in 2019 and 7 episodes of the Game of Thrones (GoT). We will release these datasets and code upon publication. 

\begin{itemize}[topsep=0.5pt]

\item We develop an approach that uses contemporaneous Twitter commentary about a televised event and the people involved to segment the event into sequential scenes.

\item We present a novel approach that uses weak supervision data and incremental learning to effectively recognize faces in video frames. Our method outperforms previous weakly supervised approaches on the task of face recognition. %while potentially robust to racial biases. 
We use this method to select video frames for the identified scenes, thus creating multimedia summaries of the event. 

\item We collect two new datasets to test our approach.  They include video frames and tweets corresponding to night 1 of the first United States (US) Democratic party presidential debates in 2019 and 7 episodes of the Game of Thrones (GoT). We will release these datasets and code upon publication. 
\end{itemize}

%\section{Discussion}

%The image selected by our multimedia summary model to indicate a sub-event had a higher accuracy than the key frame from Amazon.

%among others before clustering the faces and labeling the clusters with the improved labels. 

\noindent
\begin{center}
\textbf{Related Work}
\end{center}

%\section{Related Work}
%In this paper, we are 
This paper addresses the task of multimedia summarization of events using tweets and videos. %This task reflects the reality of live events. When a live event happens, often the first updates we get are videos of the event happening and evolving reactions in UGCs such as Twitter. Automatic multimedia summarization of such live events that is large-scale, spontaneous and unmanaged by officials within institutional settings can provide important nonscripted insights and diverse perspectives into the event in real time \cite{livingston2003gatekeeping}. 
Although previous works have addressed the  summarization of videos \cite{wang2012event,ajmal2012video,zhang2016video} and Twitter data \cite{huang2018event} separately, ours is the first to address them together.

Past work determined %has shown 
that televised events often garner considerable attention from the public and that Twitter captures large volumes of discussions and messages related to these events, in real-time ~\cite{tumasjan2010predicting, starbird2012will,paul2011you, sakaki2010earthquake, guo2013link, sakaki2010earthquake,andy2020resolving}.  %Identifying important moments during an event from social media data has been shown to be important for summarizing  \cite{andy2019winter} and gaining insights about these events %about these events \cite{shamma2009tweet}, and for capturing user reactions to these events \cite{shamma2011peaks}. 
To detect important moments of events from social media data, prior work focused on identifying the highlights of these events by using the increase in the volume of published tweets around these events \cite{nichols2012summarizing, gillani2017post, shamma2009tweet}. %; specifically, a potential highlight was selected if it had a slope that exceeded a threshold that is based on the median \cite{nichols2012summarizing} or the mean and standard deviation of the tweet volume or the change in the volume \cite{gillani2017post, shamma2009tweet}.

%. In a tweet stream, the volume of tweets may increase for a few minutes and have multiple local peaks. Since some sub-events generate less traffic than others, previous works have selected a potential highlight as a highlight if it had a slope that exceeded a threshold that is based on the median \cite{nichols2012summarizing} or the mean and standard deviation of the tweet volume or the change in the volume \cite{gillani2017post, shamma2009tweet}. 

The challenge with the moment detection models described in prior work is that some scenes are not as exciting as other scenes and thereby do not attract as many tweets. 
Our approach is different from the previous work in that it identifies the character mentions per minute and based on the frequency of character mentions, it determines a change in scenes. 
Even though the scene does not generate a lot of tweets, our approach is able to detect the scene.

Some work has been done to capture visual context (video) as well as dialogue exchanges among multiple speakers \cite{barbieri2017towards,pasunuru2018game} and use Twitter feeds related to live coverage of events to annotate the sentiment of videos associated with the live coverage \cite{sinha2014sentiment}. %In \cite{pasunuru2018game}, using a dataset based on live soccer games from  `Twitch.tv` (which is a live-video streaming service) with many users discussing about the game, in real-time, dialogue models were developed to produce language which is relevant to the live video. \cite{barbieri2017towards} proposed two tasks specific to the Twitch platform i.e. emote detection and trolling detection.
%they proposed two tasks specific to the Twitch platform: (1) Emote prediction, which predicts the emote a user would use given a set of messages, and (2) Trolling detection, which detects emotes used in trolling messages on the Twitch platform. 
Related to our work, there have also been previous approaches on multimodal summarization. In \cite{zhu2018msmo,zhu2020multimodal}, known news events discussed by news articles and images collected from these news articles are summarized. In \cite{li2018read}, a multimodal method for generating text summaries from asynchronous documents, images, audio recordings, and video of the same news event is proposed. These previous approaches assume the domain of news text, while ours assume the domain of UGCs which are shorter, noisier, often informal, idiomatic and sentiment-laden, which are excellent for summarizing real time reactions about the event; but on which summarization methods trained on clean text such as news have been observed to not work well \cite{jing2003summarization,meechan2019effect}. These approaches also assume that the event is known, while we assume a live  event that is unfolding and consisting of multiple yet-unknown scenes that first need to be detected before summarized. Further, previous multimodal approaches summarize from multiple sources and modalities that are all narration of the event. In our case, although the UGCs contain narration of the event, the videos contain dialogues which are part of, but do not narrate the event. There is a bigger semantic gap between the UGCs and videos. For example, transcription of videos in our case cannot be directly used for summaries unlike in \cite{li2017multi}, where sentences from video transcriptions are added to the summary to improve informativeness. Lastly, although previous multimodal summarization approaches assume asynchronous sources, the images are still found within news text and the videos are accompanied with narration about the same news event. In our case, there is no such obvious alignment between video and text. Instead, we make use of temporal information for aligning the reaction in tweets with the video being broadcasted at the same time frame. To alleviate the problem of time lag between the video and the reactions in tweets, we also use the change in the frequency of character mentions in tweets with the characters shown in the video to align tweets and video frames automatically and produce a multimodal timeline summary of scenes. 

With regards to face recognition, the state-of-the art in face recognition trains a neural net in a supervised manner to push the embeddings of faces from the same person near to each other, and those of different people far apart \citep{SchroffKP15}. We use these pre-trained face embeddings as feature vectors to represent faces we extract from video frames. However, different from these supervised methods that employ manually annotated faces, we use textual annotation of the show in the form of subtitles and transcripts to generate weak labels for faces. These have been shown to be useful for identifying character faces in TV shows \citep{tapaswi2015improved,miech2017learning,everingham2009taking,everingham2006hello}. Previous works add other signals such as face tracking \cite{Everingham06a,parkhi2015s}, coreference resolution from video description \cite{ramanathan2014linking}, lip movement detection \cite{tapaswi2015improved}, person tracking through face recognition, clothing appearance, speaker recognition, and contextual constraints \cite{tapaswi2012knock}, textual contents of the subtitles \cite{azab2018speaker}, activity recognition\cite{bojanowski2013finding}, and freely available image resources in the web \cite{nagrani2018benedict}. In our work, we have not used other information outside of subtitles and transcripts, which we can pursue in future work. Instead, we have concentrated on improving the learning with weak labels using different loss functions and incremental strategies of relabeling ambiguous labels. Contrary from previous works that assume all data from all the episodes to be available at the beginning of training, we perform multiple stage learning where we expose our system to training data from more episodes at each stage.

Many previous works on face recognition with weak/no labels have used EM-like algorithms \cite{franc2017learning}, clustering algorithms \cite{sharma2019self} (we use k-means clustering as one of our baselines), and multi-instance learning (MIL) algorithms that require bags of faces during train and test time \cite{haurilet2016naming}. In our case, these bags of faces are not given during test time. Other work has learned to generate descriptions and jointly ground mentioned characters in videos \cite{rohrbach2017generating}. In contrast, we learn to detect scenes from tweets and describe them using descriptions from tweets and frames from the video.

\begin{figure*}[t]
\includegraphics[scale=0.275]{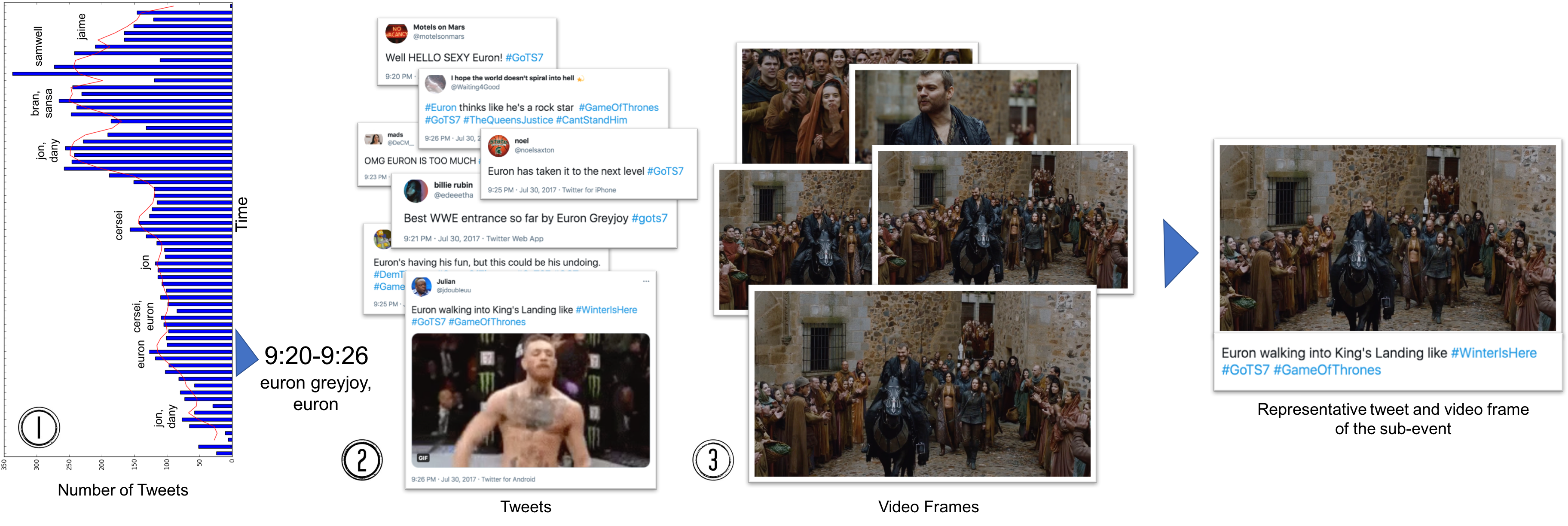}
\centering
		\caption{Overview of our approach that consists of three steps: (1) identifies scenes, its time period and the characters/persons mentioned in it, (2) selects the most representative tweet among the tweets in that time period that mention the characters in the scene, (3) selects the most representative frame among video frames around that time period that contain the characters' faces. }
		\label{overview}
		\vspace{-1em}
\end{figure*}

\noindent
\begin{center}
\textbf{Task Definition}
\end{center}
\noindent
The goal in this work is as follows: given a televised event such as a Presidential debate, use Twitter data related to the event (published while the event is airing) and video of the event to create a multimedia summary of the event.

\noindent
\begin{center}
\textbf{Our Approach}
\end{center}
\noindent

%\section{Our Approach}

In this section, we describe our approach to this task. Our approach consists of 3 stages i.e. (1) identify sequential scenes (2) select tweet indicative of each scene, and (3) select video frames to represent each indicative tweet. Figure \ref{overview} shows an overview of our approach.\\

\begin{algorithm}[h]
\begin{flushleft}
\noindent \textbf{Input}: Tweets, \textbf{\textit{T}} (related to a televised event) published while an  event is ongoing 

\noindent \textbf{Output}: All sequential scenes in the event from the tweets
\end{flushleft}
\caption{Scene identification from Twitter data}
\label{array-sum}
\begin{algorithmic}[1]

\Procedure{SceneDetection}{\textit{T}}
      
   \For {tweets in \textit{T} published in a minute}
         \State \textit{create a bin to store these tweets} 
   \EndFor
    \For {each bin identify the (\%) of each character mentions}
         \If {\textit{(\%)  $>$ k\% and character not from previous scene i.e., mentioned less than m\% time in previous  scene}}
            \State \textit{mark as start of new scene}
            
        \EndIf 
    \EndFor 
   
	\State Return SceneChange(\textit{minute of scene change and characters/persons})
\EndProcedure
\end{algorithmic}
\end{algorithm}

\noindent
\textbf{Stage 1 - Identifying sequential scenes:}
%\subsection{Stage 1 - Identifying sequential scenes} %Person Mentions and Scenes}
\label{charid}
The first step in our pipeline is to tag mentions of people or characters. In televised events, these are often known prior to the start of the event. Nevertheless tagging known people is still a challenging task, since tweets often refer to characters using nicknames, actors' names, or other aliases.  For each character in an event, we construct an alias list. %, following \newcite{andy2017constructing}. 
Our alias lists consist of their first names, last names, and the nickname listed in the first paragraph of the character's Wikipedia page. %For each character, \textit{a} in an event, we construct an alias list: \textit{a\textsubscript{1}, a\textsubscript{2}, ..., a\textsubscript{n}}, where \textit{a\textsubscript{i}} %. , following \newcite{andy2017constructing}. Our alias lists consists of their first names, last names, and the nickname listed in the first paragraph of the character's Wikipedia page. 
For example, the the alias list for the Presidential candidate \textit{Beto O'Rourke} is \textit{Robert} (his given name), \textit{O'Rourke}, and \textit{Beto} (his nickname).  The alias list for the GoT character, \textit{Petyr Baelish} is \textit{Petyr}, \textit{Baelish}, and \textit{Littlefinger}.

%\subsection{Stage 2: Detecting Sub-Events}

%Events such as Presidential debates and GoT have a tendency for unanticipated/unexpected moments  such as a candidate calling out another candidate for misrepresenting a point or candidates talking over each other when trying to make a point. Also, some sub-events are exciting and people publish a lot of tweets around these sub-events, however, some other sub-events are not as exciting, hence generating fewer published tweets.  We observed that if an exciting sub-event is followed by another sub-event that is not as exciting, the characters in the exciting sub-event will have more mentions in tweets published during the time period of the new sub-event. For example, on night 2 of the first 2019 US Democratic party Presidential debate when Senator Kamala Harris and Vice-President Joe Biden had a heated exchange, there was an increase in the number of published tweets during this time period and  a lot of these tweets mentioned either Kamala Harris, Joe Biden, or both candidates. The next candidate to speak after this interaction  was Senator Bennie Sanders. Although Bernie Sanders had a lot of mentions in tweets published during this time period, Kamala Harris and Joe Biden were both mentioned more times in tweets published during this time period. 

To identify scenes, our approach collects then groups tweets about an event chronologically into one minute bins.  
For each character and each bin,  the percentage of tweets in the bin that mention the character is calculated.  Our scene detection approach (Algorithm \ref{array-sum}) 
chooses when to create a new scene based on changes in character mentions.
A new scene is triggered if a character bin spikes: that is if the fraction of tweets where the character is mentioned in the current minute exceeds a hyperparameter $k$, while the fraction of tweets mentioning them in the previously determined scene is lower than a hyperparameter $m$.
The values of $k$ and $m$ can be optimized on a development set. 
Our approach outputs the start times of scenes and the character(s) that determined the change. The start time of a new scene also signifies the end time of the previous scene.

%For example, during night 1 of the first 2019 Democratic debate when Kamala Harris and Joe Biden had a heated exchange, of the \textbf{N} of tweets published during this time period, \textbf{Y} mentioned Kamala Harris alone, \textbf{Z} mentioned Joe Biden alone; however\textbf{D} mentioned both Kamala Harris and Joe Biden.
 
%We varied this percentage and observed that 25\% was the optimal percentage. Figure \ref{scene} shows an outline of our scene detection algorithm

%\section{Our Algorithm to Select Video Frames for Sub-Events}

%To provide a complete picture of the predicted sub-events from tweets, our algorithm has 2 stages: in the first stage, for each predicted sub-event, we first select the tweet from its predicted time period that best describes the sub-event. In the second stage, using the selected tweet, we select video frames that contain faces of characters mentioned in the tweet. Together, the tweet and the video frames provide a complete depiction: description and picture of the sub-event. 
\noindent
\textbf{Stage 2 - Select tweet indicative of each scene:}
%\subsection{Stage 2 - Select tweet indicative of each scene}% Tweet Selection}
To find the tweet that best describes each scene, we use BERT \cite{devlin2018bert} to create embeddings for each tweet by averaging its token embeddings. For the period of a predicted scene, we compute the centroid of the tweets and select the tweet whose embedding is closest, in terms of cosine similarity, to the centroid \textit{and} which contains the names of characters that spiked during the scene.  Our approach selects this tweet as the description of the scene.

%Intuitively, by selecting the tweet that is closest to the centroid and contains the characters of the sub-event, we'll be selecting the tweet that is most similar to all the other tweets published in that period and that talks about characters that are involved in the sub-event. Some examples of the tweet descriptions of sub-events we identify in GoT and their corresponding descriptions from Amazon Prime are shown in Figure \ref{examples}. 
\noindent
\textbf{Stage 3 - Select video frames to represent each indicative tweet:}
%\subsection{Stage 3 - Select video frames to represent each indicative tweet}
Once a tweet is selected to describe a scene, our approach selects frames from videos of the event to visually represent this scene. Since our scene identification is driven by people/character mentions, we select frames from the video of the scene which show the people mentioned in the tweet describing the scene. To do so, we train a model that recognizes faces in the frames: given a face \textit{x}, the model produces its label \textit{y} i.e., the person's name. We describe our face recognition model in detail in  \textbf{Section Face Recognition Model}.%\ref{facerecogmodel}. 

For each identified scene and its tweet description, our approach uses our trained face recognition model to select video frames from that time period that contain faces of the characters mentioned in the tweet.
%Specifically, for each episode, we process identified sub-events from that episode sequentially from first to last. For a sub-event from episode $e$ that has the characters $Z=\{y_i \in Y_\text{tot}\}$ mentioned in its tweet description, we find the sequence of video frames from $e$ to align with the sub-event. The alignment starts from the frame that contains the face of any character in $Z$ and is not part of the sequence aligned to the previous sub-event, and ends with the starting frame of the next sub-event. 

Similar to how we find the best tweet for each scene, for each character in the scene, we find the centroid of the frames that contain the character mentioned in the tweet. We use ResNet50 \cite{he2016deep} to obtain vector representations of the frames and select the frame that is closest to the centroid \textit{and} has the highest confidence of containing the character. We also select the frame that contains all the characters in the same way. Intuitively, by selecting the frame that is closest to the centroid and contains the characters of the scene, we will be selecting the frame that is most similar to all the other frames published in that period and that contains the characters involved in the scene. 
%If the sub-event contains more than one characters, for each character, we select a frame that fits the requirement above. Additionally, we also select a frame that is closest to the centroid and contains all the characters with the highest confidence. 
We order the selected frames based on their time stamps and return this sequence of frames to visually represent the scene.

\noindent
\textbf{Face Recognition Model:}
%\subsection{Face Recognition Model}
\label{facerecogmodel}
Using video subtitles and transcripts, we can label our video frame with \textit{weak} labels for the faces shown in the frame---that is, the face labels are not fully specified since subtitles and transcripts only give us the information of who is \textit{speaking}, but not necessarily who \textit{appears} in the frame or \textit{who is who} in the case of multiple faces in the frame. 
% . These indicate who speaks in a given scene but the 
For every detected face in the frame, a character's identity is hence  ambiguous: each face is partially labeled with a set of characters speaking in the frame. In our dataset, $\sim$76\% of faces in GoT and $\sim$73\% of faces in the debate are labeled with multiple names (see \textbf{Datasets}). 
Our goal is to learn a face recognition model that can refine and disambiguate the labels of the training faces and also generalize to unseen data. In contrast to the standard supervised setting where each training face is labeled with an unambiguous single label, in our dataset each training face \textit{x} is labeled with a set of possible labels $Y$, only one of which is correct. $Y$ is a subset of $Y_\text{tot}$: the set of all possible characters in the video.

In a fully-supervised multiclass setting, where (\textit{x}, \textit{y}) pairs are given, we can learn our model parameters $\theta$ by minimizing the negative log likelihood of $y$ given the input $x$ with respect to $\theta$:  $\mathcal{L}_\text{sup}(\theta) = -\text{log}\ \text{P}(y|x;\theta)$. In our partially-labeled training scenario, the model has access to \textit{x} and \textit{Y} = \{$y_1$, $y_2$, ..., $y_n$\}. %We explore several works that addressed this partially labeled problem. 
%Past research has used the expectation-maximization (EM) like algorithms, minimizing the (negative) marginal likelihood (MML) to estimate $\theta$ and the true label: $\mathcal{L}_\text{MML}(\theta) = -\text{log}\ \sum\limits_{y_i \in Y}  \text{P}(y_i|x;\theta)$. 
%
%
%Most recently \citet{min2019discrete} develops a variant of MML called hard EM learning scheme that computes gradients relative to the most likely solution at each update: 
%$$\mathcal{L}_\text{hardEM}(\theta) = -\text{log}\ \max\limits_{y_i \in Y}  \text{P}(y_i|x;\theta)$$

%We experiment with an approach that addresses the problem of MML assigning high probabilities to any subset of \textit{Y}, which is not desirable in our case; since only one \textit{y} in \textit{Y} is correct others should be ideally assigned very low probability. This hardEM approach has been shown to achieve state-of-the-art performance on QA tasks where each question is accompanied with a set of possible solutions that contains one correct option.

We experiment with a hard expectation-maximization (EM) algorithm from  \cite{min2019discrete}, which minimizes the (negative) marginal likelihood, while attempting to assign a high probability to only one label:
$$\mathcal{L}_\text{hardEM}(\theta) = -\text{log}\ \max\limits_{y_i \in Y}  \text{P}(y_i|x;\theta)$$

We also experiment with an average categorical cross entropy (aveCE) loss  \cite{mahajan2018exploring,joulin2016learning,cour2009learning}: 
$$\mathcal{L}_\text{ce}(\theta) = -\frac{1}{|Y|} \sum\limits_{y_i \in Y} \text{log}\  \text{P}(y_i|x;\theta)$$
where $\text{P}(y_i|x;\theta)$ is a softmax function $\frac{e^{\theta_i x}}{\sum_{y_k \in Y_\text{tot}} e^{\theta_k x}}$ and $\theta_i$ is the parameters of the model for character $y_i$. We use stochastic gradient descent to learn parameters to minimize the loss. 

If the set \textit{Y} contains a single label \textit{y}, then this loss reduces to the regular multiclass cross entropy loss. However, when \textit{Y} is not a singleton, this loss will drive up the average of the scores of the labels in \textit{Y}. Intuitively this will mean that if the score of the correct label is large enough, the other labels in the set do not need to be positive. This loss has been shown to work well for some multi-label classification problems \cite{mahajan2018exploring}. In this work, we explore its use for our partial-label problem. 

We employ different strategies for training our face recognition model: (1) by training the model with all our training data at once or (2) incrementally by sequentially exposing our model to training data from one episode at a time. 

In the incremental strategy, we retrain our model after each episode with \textit{all} the data it has seen so far. We then used the trained model to explicitly disambiguate each face \textit{x} in our training data with its current most likely solutions \textit{y} $\in Y$. 
%For this ``relabeling", as our model in the incremental setting is not exposed to all the training faces at once rather, to fewer faces at a time, we use the idea similar to Prototypical Networks \cite{snell2017prototypical}, which has been shown to be effective for few-shot incremental learning. %, to disambiguate the weakly supervised faces in our training data.
Specifically, similar to the idea of Prototypical Networks \cite{snell2017prototypical} that has been shown effective for incremental learning, we use the model to compute a centroid (prototype) for each character label: the average embedding of faces assigned by the model to the label; and relabel the faces based on the distance to each prototype. This relabeled data is then used to retrain the model in the next iteration. 

For the debate dataset, since we only have one ``episode" (night 1 of the first debate), we train our model for several iterations on this ``episode" with relabeling.

\noindent
\begin{center}
\textbf{Datasets}
\end{center}
\noindent

%\section{Datasets}

We demonstrate our approach on data from two types of televised events: presidential debates and a TV show. %Scenes vary depending on the type of event; for example in a presidential debate, an example of a scene is a candidate responding to a question posed by the moderator, however, in a TV show, a scene could be an interaction between two characters.

	\begin{table*}[ht]

\centering

\begin{tabular}{>{\centering\arraybackslash}m{2.0cm}|>{\centering\arraybackslash}m{2.0cm}|>{\centering\arraybackslash}m{11.0cm}}
\hline
\textbf{Time} & \textbf{Candidate} & \textbf{New York Time reporters' live chats} \\
\hline
9:11 PM & Julian Castro  &  "Castro gets the next question, about the pay gap and what he would do to ensure women are paid fairly. He says he wants to pass the Equal Rights Amendment and gets some loud cheers from the audience." \\
\hline
9:23 PM  &Beto O'Rouke & "O’Rourke is asked why he no longer wants to abolish private insurance. He explains that if you’re a member of a union, you should be able to keep your health care plan. He says directly that he would not get rid of private insurance."\\
\hline

\end{tabular}
\caption{Example of New York Times reporters' live chats we use as our ground truth scenes for the Presidential Debate event.}

\label{tab:New_York_Times_posts}
\vspace{-1.5em}
\end{table*}

\noindent
\textbf{Presidential Debate Dataset:}
%\subsection{Presidential Debate Dataset}
The first 2019 US Democratic party Presidential debate was held on two nights. We collected tweets and video of night 1 of the debate. Night 1 of the debate had 10 candidates debating 
with 5 moderators.
Similar to prior works \cite{andy2019winter,andy2020resolving},
using the Twitter streaming API, we collected over 50,000
timestamped and temporally ordered tweets and re-tweets that mentioned the word \textit{``debate"} during the time periods in which night 1 of the debates aired.
For ground truth scenes, we segmented the debate based on when each Presidential candidate spoke. We obtained these from the New York Times (NYT) reporters' live chats\footnote{\url{https://www.NYT.com/interactive/2019/06/26/us/politics/democratic-debate-live-chat.html}} about the debates. These live chats include the timestamp in which each candidate spoke and a brief description of what the candidate spoke about as shown in  Table \ref{tab:New_York_Times_posts}.  
One of the co-authors reviewed the NYT reporters live chats and marked the timestamp each candidate started speaking as the beginning of a new scene; the timestamp of a new scene also indicates the end time of the previous scene.

%the timestamp a different person started speaking was marked the beginning of a new scene and the end of the previous scene.  

%\textbf{Derry TODO: Write about extracted faces for the debates}

\noindent
\textbf{Game of Thrones Dataset:}
%\subsection{Game of Thrones Dataset}
For our Game of Thrones (GoT) dataset, we collected tweets for all the 7 episodes of GoT season 7. Each episode lasted for approximately an hour. Similar to prior works \cite{andy2019winter,andy2020resolving}, we used the Twitter streaming API to collect time-stamped and temporally ordered tweets containing the ``\#gots7", a popular hashtag for the show, while each episode aired.  We collected over 87,000 tweets (averaging 12,439 per episode).  %Figure \ref{fig:spike_1} shows an example histogram for one episode. 
Character names and alias were collected from Wikipedia.  
For ground-truth scenes, we used the scene timestamps from the Amazon Prime video streaming service DVD chapter summaries. The DVD chapter summaries of GoT also contain a short description of each scene, and an image key frame for each scene. 

\noindent
\textbf{Video Frames and Faces:}
%\subsection{Video Frames and Faces}
For each of these datasets, in addition to the Twitter data, we extracted faces for the debate and for each GoT episode. We first sample frames, once per second, from the videos of the debate and of each GoT episode.  Then, we extract faces from each frame with the OpenFace library \cite{amos2016openface}, and obtain embeddings of the faces using the pre-trained FaceNet embeddings \cite{SchroffKP15}. We discard faces from frames that contain more than 5 faces. In total, using OpenFace we obtained 10,318 faces from GoT episodes (averaging 1,474 faces per episode) and 9,979 faces from the debate. 
% and train only on extracted faces that have size equal to or more than 10 KB (to reduce existence of blurry faces in our training data). 

We obtain weak labels for the faces via subtitles (which record \textit{what} is said and \textit{when}, but not by \textit{whom}) and transcripts (which record \textit{who} says \textit{what} but not \textit{when}), following previous work on face recognition in TV-Shows \cite{everingham2009taking}.  By matching \textit{what} is said in subtitles and transcripts, we create labels of a face based on \textit{who} is speaking \textit{when} the face is shown. However, knowledge that a character is speaking gives only a very weak cue that the character is in the frame. The speaking character may not always be visible and other characters who are not speaking or who speak before or after the character may be shown. Hence, we use all speaking character names in and within 15 seconds window of each second frame as labels. Each face in our dataset may thus be labeled with multiple names. On average, for each episode of GoT, $\sim$76\% of the faces are labeled with multiple names (a total of 7,895 faces). For the debate, $\sim$73\% of the faces are labeled with multiple names (a total of 7,301 faces). 

%as discussed in Section \ref{facerecogmodel}. %(Figure~\ref{motivation}).

Since transcripts (\textit{who} says \textit{what}) are only available for episode 1-6 of GoT season 7, to obtain weak face labels for the remaining episode i.e., episode 7, we train a deep speaker recognition model (a 3-layer feed forward neural network) on speech\footnote{We use pre-trained speech representation provided by https://www.voicebiometry.org/} from episode 1-6 that are labeled with character names by matching the speech time stamps and \textit{when} characters are speaking from the subtitles and transcripts of these episodes. Our speaker recognition model achieves 96.17\% 10-fold cross validation accuracy for recognizing speaker from speech in training. We use this model to predict \textit{who} speaks \textit{when} in episode 7 and use these predictions to obtain weak labels for faces in this episode.

\noindent
\begin{center}
\textbf{Evaluation}
\end{center}

%\section{Evaluation}

Similar to prior work \cite{bekoulis2019sub}, %our approach's predicted scene boundaries are evaluated by comparing the predicted timestamp boundaries against the ground-truth data i.e. 
a predicted scene is considered correct if the predicted start time is within the timestamp boundary of the ground-truth reference. We compute precision, recall, and F1 scores of our approach for identifying scenes. In the future, to evaluate the scene detection algorithm, we will set a margin of error %for the scene detection algorithm; 
so if the predicted start time of a scene is between the ground truth start time and the margin of error, the predicted time is considered correct otherwise it is not correct.

We evaluate our face recognition model on our presidential debate and GoT datasets.  %\textbf{For the Presidential debate, we created a test set TODO: Derry.....}. 
For the presidential debate dataset, we create a test set with 113 randomly sampled faces, which we manually annotated with 15 labels (the candidates and moderator names). We use the rest of the $\sim$9k faces for training. Similarly, for, GoT we created a test set with 1,079 randomly sampled faces from episode 7, which we manually annotated with 45 labels (the character names). We use the rest of the faces from GoT season 7 ($\sim$11k faces) for training. We compute the micro accuracies of our face recognition on these test sets (\textbf{Section Face Recognition Results}). %(Section \ref{facerecogresults}). 

For the selection of tweets and images to represent each scene, we perform  qualitative and quantitative analysis by comparing against our ground truth, gold standard, real life sources that are expertly and professionally curated (\textbf{Section Multimedia Summary Results}). %: the NYT reporters live chats of the presidential debate (which contain a short description of each scene) and the Amazon Prime DVD chapter summaries of GoT (which contain an image key frame and and a short description of each scene) (Section \textbf{Multimedia Summary Results}).\\ %\ref{multimediaresults}).\\ 

%Amazon Prime sub-event descriptions for GoT (which contain an image key frame and and a short description of each scene) and the NYT live chats for the presidential debate (which contain a short description of each scene).

%\begin{figure}
%\includegraphics[scale=0.16]{episode_4_new_new_smaller.png}
%		\caption{Histogram of number of tweets published per minute during an episode of GoT season 7.
%			The red line, which forms peaks, shows the mean of the number of tweets published every 3 minutes.
%			The names of the character that had the most mentions during each peak in tweet volume are also shown.}
%		\label{fig:spike_1}
%\end{figure}
\noindent
\textbf{Scene Identification Results:}
%\subsection{Sub-event Identification Results}
\label{subeventresults} From the ground truth data (i.e. NYT reporters' live chats and Amazon prime video streaming service DVD chapter summaries), there were 47 scenes in the presidential debate and 90 scenes in GoT.

%Our model identified 26 out of 47 scenes in the presidential debate dataset and 76 out of 90 scenes in the GoT dataset.

%(in the discussions section, we discuss this result). 
%On the average, the Presidential debate scenes lasted for 1.3 minutes and the GoT scenes lasted for 4 minutes.

%each scene in the Presidential debates had lasted for 1.3 minutes also, for GoT, the average length of a scene was 4 minutes.

%. On the average 431 tweets were associated with the time period of scenes published during the time period of the scene and for GoT\textbf{TODO: Y} tweets on average were published during the time period of the scenes.

%\textbf{TODO: X} 
%and 76 out of 90 scenes in the presidential debate and GoT, respectively. 

%\textbf{TOD: state how many scenes were detected and evaluated in GoT and the debates}

%of each episode of GoT and the Presidential debate,we plot the number of tweets that were published at each minute of an episode of GoT and the Presidential debates, respectively. Since data at the minute level is quite noisy and to smooth out short-term fluctuations, we calculated the the mean of the number of tweets published every 3 minutes which forms peaks in the tweet volume, as shown in figure \ref{fig:spike_1}. 

We compare our scene identification approach to two baselines:

\noindent\textbf{Baseline 1:} %Spikes in the volume of tweets only. 
This baseline identifies a scene if there is a peak in the volume of tweets and records the start time of the ascent of the peak and the time of the descent of the peak as the start and end times of the scene.

%\paragraph
\noindent\textbf{Baseline 2:}  This spike-based baseline uses the moment identification model from \cite{shamma2009tweet, gillani2017post}, which uses the mean and standard deviation to determine the threshold for selecting scenes.

Tables \ref{tab:presidentialdebate} and \ref{tab:GoT-time} show that our entity-spike approach for identifying scenes outperforms these two baselines for identifying scenes in terms of precision, recall, and F1 scores when evaluated against %both 
the ground truth.% NYT reporters' reported sub-events of the presidential debate and the DVD chapter sub-events of GoT.

\begin{table}[h]
	%\small
\centering
\begin{tabular}{|l|l|l|l|}%{lllll}
\hline
 \centering

 \textbf{Approaches} & \textbf{Precision} & \textbf{Recall} & \textbf{F1}\\
 \hline
  \textbf{Our Model}  & \textbf{0.79} & \textbf{0.71} & \textbf{0.74}\\
  Baseline 1 & 0.77& 0.55 & 0.64 \\
  Baseline 2  & 0.75& 0.37& 0.49\\
 % Baseline 3  &  &  & \\
  \hline
  %Context with Entity  & 0.65 & 0.63 & \\
  %No Entity Count  & 0.64 & 0.74 & 0.6\\
  %GOT 4 highlight 5  & 4 & 0.75 & 4\\

 %   Soccer game & \ 0.72
   % JayZ & Music Beats, Young Joc\\

\end{tabular}
\caption{Our approach outperforms the baselines for identifying scenes around the Presidential debate on the NYT ground truth scenes}
\label{tab:presidentialdebate}
  \vspace{-1.5em}
\end{table}

\begin{table}[h]

\centering
%\small
\begin{tabular}{|l|l|l|l|}%{lllll}
\hline
 \centering

 \textbf{Approaches} & \textbf{Precision} & \textbf{Recall} & \textbf{F1}\\
 \hline
  \textbf{Our Model}  & \textbf{0.80} & \textbf{0.58} & \textbf{0.67}\\
  Baseline 1 & 0.77& 0.35 &0.48  \\
  Baseline 2  & 0.58&0.30 & 0.39\\

  \hline
\end{tabular}
\caption{Our approach outperforms the baselines for identifying scenes of GoT season 7 on the Amazon Prime DVD chapter ground truth scenes}
  \vspace{-1em}
\label{tab:GoT-time}
\end{table}

%\paragraph{}
%\begin{table}[H]
%\centering
%\begin{tabular}{|l|l|l|l|}%{lllll}
%\hline
% \centering

% Threshold & Precision & Recall & F1\\
% \hline
%  5  & 0.67 & 0.78 & 0.72  \\
%  \textbf{6}  & \textbf{0.72}& \textbf{0.73} & \textbf{0.72} \\
%  7  & 0.69& 0.72 & 0.70 \\
%  8  & 0.74& 0.65 & 0.69\\
%  9  & 0.77& 0.61 & 0.68\\
%  10  & 0.81& 0.49& 0.61 \\
 
%  \hline

%\end{tabular}
%\caption{Result from applying our model to subtitiles in GoT}
%\label{tab:subtitles}
%\end{table}

% Siyi added here:
% To find the best tweet that describes our predicted sub-event the most, we first tokenize and compute the sentence embedding for each tweet sent during the time in that sub-event using Pre-trained base uncased Bert model, and then find the centroid of these embeddings by taking the average of them, and then find the tweet that has all the names that spike in this scene and is most cosine-similar to this centroid. We take this tweet as the most descriptive tweet of the scene since we believe it 

%Table \ref{tab:summary} shows some of the sub-events - in each episode of GOTS7, that our algorithm identified and the baselines did not. Our algorithm outperformed all the baseline as shown in tables \ref{tab:dev} and \ref{tab:presidentialdebate}.  

  \begin{table}
	\centering
	%\small
	\begin{tabular}{|l|c|c|}
		\hline 
		\textbf{Method} & \textbf{GoT} & \textbf{Debate}%all &  %\multicolumn{3}{c|}{ %\textbf{Annotations Source}} \\
		%&\textsc{Twt} & \textsc{Sub} & \textsc{Comb} 
		\\
		\hline
		Incremental + aveCE + relabeling & \textbf{85.1} & \textbf{94.7}\\
		Incremental + aveCE & 77.8 & 92.0 \\
		Incremental + hardEM + relabeling & 83.1 & 90.3\\
		Incremental + hardEM  & 79.5 & 90.3\\
		All-at-once + aveCE & 78.3& 92.0 \\
		All-at-once + hardEM & 80.7 & 91.2 \\
		k-Means & 71.5 & 87.1 \\
		Logistic Regression & 61.2 & 88.5 \\
		FFNN + multilabel loss &59.5 & 87.6\\
		FFNN + cross-entropy loss &45.2 & 85.0\\ 
		%Best of KMeans on Test (K=49) & 64.9 & \% \\
		%Best of Spectral on Test (K=60) & 61.6\%\\
		%Label-Aware KMeans & 52.5\%\\
		%Speaker Recognition Labels & 21.9\%\\
		\hline
	\end{tabular}
	\caption{Accuracies of labeling faces in the face recognition test sets. Our approach outperforms baselines and other previous methods with weak labels on both GoT and Debate face recognition test sets.} %from tweet, subtitles, or their combination} %(highest accuracy in bold)}
	  \vspace{-1.5em}
	\label{facerecog}
	\end{table}

%	\begin{figure}
%		\includegraphics[width=0.45\textwidth]{faces.png}
%		\caption{The top-10 faces predicted by our model to be the character \textit{Hotpie} after \textit{watching} upto episode 2 (ep2), upto episode 4 (ep4), and upto episode 7 (ep7) of GoT. The images show how our model learns to disambiguate \textit{Hotpie} better as it sees more episodes. As \textit{Hotpie} appears only once in GoT season 7; in a scene with \textit{Arya} in episode 2, his top-10 in this episode contain \textit{Arya}'s faces (marked). However as the model sees more episodes (ep4, ep7), it is able to recognize and disambiguate \textit{Hotpie} from \textit{Arya}.}
%		\label{ot}
%	\end{figure}

\noindent
\textbf{Face Recognition Results:}
%\subsection{Face Recognition Results}
\label{facerecogresults}
Table \ref{facerecog} gives the micro accuracies of our face recognition models. We compare the performance of our models with baselines such as k-Means clustering, Logistic Regression, and a three-layer feed-forward neural network (FFNN). For k-means, we first cluster the training and test faces based on their embeddings with k=45 for GoT and k=15 for the debate dataset. Then, for each cluster, we label the faces in the cluster with its majority class and compute the accuracy on test. 

We observe that our incremental strategy with average CE loss and explicit relabeling performs the best on the test set. We also find that the typical loss function used for learning such as the cross-entropy or multilabel loss struggles to learn given the weak and ambiguous labels for the faces in our dataset (Table \ref{facerecog}). %We find that explicit relabeling improves the performance of the CE model.

We also observe that incremental learning with relabeling improves performance as it allows the model to better disambiguate faces over time. %For example, the character \textit{Hotpie} appears only once in the whole of GoT season 7 (in episode 2), and with another character \textit{Arya}. A model that only learns from ambiguous labels up to this episode may mistake his face with \textit{Arya}'s. However, a model that continues to learn from more episodes will learn to disambiguate his and \textit{Arya}'s faces better as it continues to see \textit{Arya} separately and/or with other characters in other frames (Figure \ref{ot}). 
We also observe the same with other characters that appear in the same frames frequently (e.g., \textit{Jaime} and \textit{Cersei}, or \textit{Jon} and \textit{Sansa}). In early episodes, the model cannot disambiguate \textit{Jaime} from \textit{Cersei} (or \textit{Jon} from \textit{Sansa}) well as these characters always appear in scenes together. However, as the model continues to see them in other frames, it learns to disambiguate these characters.% (%t-SNE plots \cite{maaten2008visualizing} of character embeddings and predicted labels (colors) are shown in 

%Figure \ref{change}).

%		\begin{figure}
%		\centering
%		\includegraphics[width=0.30\textwidth]{change.png}
%		\caption{The t-SNE plots \cite{maaten2008visualizing} of characters' face embeddings and their predicted labels (colors). In the early episode (Ep2, left), the model cannot disambiguate  \textit{Cersei} (red) from \textit{Jaime} (green) (or \textit{Jon} (blue) from \textit{Sansa} (purple)) as these characters always appear together in scenes i.e., faces in the top right and bottom left are both labeled with red and green. In the later episode (Ep4, right), as the model sees more of these characters in other scenes with other people, it learns to disambiguate these characters i.e., the different color dots are better separated.}
%		\label{change}
%	\end{figure}

	%	\begin{figure}
	%	\includegraphics[width=0.3\textwidth]{frequency.png}
	%	\caption{Average accuracy across characters increases with the frequency of faces (cumulative and as a fraction of all faces) seen by the model.}
	%	\label{frequ}
		
	%\end{figure}
	
We also observe in GoT that there is a weak positive correlation (Pearson's \textit{r} = .2824) between the model's accuracy at recognizing a character and the frequency with which the character's face is shown i.e., the more frequent a character is shown, the more accurate is the model at recognizing the character. Also, the more faces the model sees, the better it is at recognizing faces; in average, accuracy across characters increases as the model is exposed to more faces. %(Figure \ref{frequ}).\\ 

%More interestingly, we find that our model, which incrementally learns to recognize faces as they are given, appears robust toward racial biases. When we compare average accuracies over white (\textit{Tyene}, \textit{Yara}, \textit{Brienne}, \textit{Melisandre}, \textit{Olenna}) and non-white (\textit{Ellaria}, \textit{Missandei}) supporting actresses in GoT\footnote{We choose supporting actresses to control for gender and screen time: none of the main actresses of GoT season 7 is a person of color.}, we observe that the average accuracy of our model at recognizing these non-white actresses is higher than its accuracy at recognizing the white actresses; and that the two accuracies increase with the frequency of faces shown to the model (Figure~\ref{robust}). In the future, it will be interesting to further explore if incremental learning can help against biases compared to supervised learning on all data, particularly when there are biases/imbalance in the data. 

%		\begin{figure}
%		\includegraphics[width=0.3\textwidth]{whiteNonWhite.png}
%		\caption{Average accuracies of our model (line graphs) at recognizing white and non-white actresses. The average accuracy is higher for non-white actresses. The average accuracies also increase with the frequencies of faces (cumulative and as fractions of all supporting actresses' faces) shown to the model (bar graphs).}
%		\label{robust}
%	\end{figure}

\begin{figure*}[h]
\includegraphics[scale=0.270]{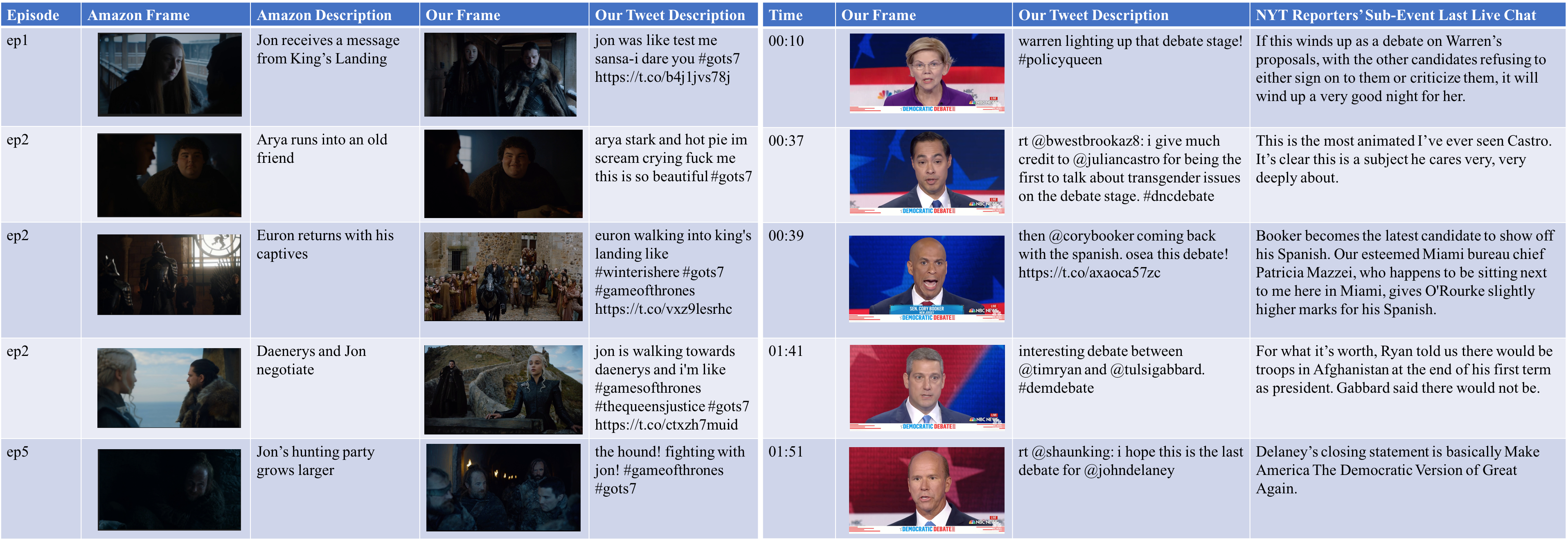}
\centering
		\caption{Multimedia summaries of GoT and Debate scenes generated automatically by our approach, compared with human expert and professional summaries of key frames and scene descriptions from Amazon Prime and live chats from NYT respectively.}%. %The name variations spike together in a given time period}
		\label{examples}
		\vspace{-1.5em}
\end{figure*}

\noindent
\textbf{Multimedia Summary Results:}
%\subsection{Multimedia Summary Results}
\label{multimediaresults}
For GoT, we compare our model's multimedia summary results against the scene descriptions and pictures from Amazon Prime DVD Chapter Summaries and show quantitative and qualitative results. For the presidential debate, since the NYT reporters live chat describes scenes but does not have images of the scenes, we compare our model's tweet summary to the last NYT reporters' live chat in each identified scene; the intuition here is that the last blog post in a scene captures the summary of what happened in that scene. Side-by-side comparison of our automatically generated multimedia summaries to Amazon DVD Chapter Summaries and NYT reporters' live chats are shown in Figure \ref{examples}.

Given that tweets are more informal and noisier compared to text from news articles or text written by experts, ROUGE scores, which are determined by comparing the overlapping n-grams between a generated summary and a reference summary is not the best evaluation metric for tweet summaries, hence, we use humans who had watched these events to evaluate our models summaries.

%To make these comparisons, we use Amazon Mechanical Turk (MTurk).

%\subsubsection{Selecting the best multimedia summary}
\noindent
\textbf{GoT:} For each identified scene, %given the Amazon key frame and the brief description of the sub-event and our multimedia summary predicted by our model, 
we ask 5 Amazon mechanical turk (MTurk) respondents - who had watched GoT, the following questions:
%the Turkers (which we determined had watched GoT) the following questions: 

\begin{enumerate}[label=(\Alph*)]%,topsep=0.5pt]
\item Given the image from the Amazon Prime DVD chapter summary and the image from our predicted multimedia summary for the scene, select the image that best depicts the scene?
%\paragraph{}
\item Given the brief description of the scene from Amazon Prime and the tweet summary from our predicted multimedia summary, select the summary that best previews the scene?
%\paragraph{}
\item In a few words, can you state why you selected the summary that you did and state what you think of the summary that was not selected (this question was marked as optional)?
%\paragraph{}
\item Given the Amazon key frame and the corresponding brief description of a scene and the multimedia summary of the scene predicted by our model, select the image \textit{and} summary pair that best  \textbf{depicts and previews the scene}?

%\item given the Amazon key frame and the corresponding brief description of a sub-event and the multimedia summary of the sub-event predicted by our model, select the image \textit{and} summary pair that best \textbf{captures the feelings and emotions associated with the scene}.
\end{enumerate}

For questions (A), (B), and (D) above, we select an image, a summary, and an image and summary pair, respectively, if 3 or more of the respondents selected it. We calculated the annotator agreement using intraclass correlation; the intraclass correlation was 0.84. %\textbf{TODO: Include inter annotator agreement}. 
The quantitative and qualitative results for the analysis on GoT are shown in Tables \ref{tab:GoT-Mturk} and \ref{tab:GoT-Mturk-qual}, respectively. We observe that, as shown in Table \ref{tab:GoT-Mturk}, respondents prefer images from our multimedia summary model 65\% of the time and prefer the summaries from our model 30\% of the time. However when presented with the image \textit{and} summary pairs from our model, some respondents who had previously preferred the summaries from Amazon Prime, selected the image and summary pairs from our model.

\begin{table}[h]

\centering
%\small
\begin{tabular}{|l|l|l|l|}%{lllll}
\hline
 \centering
 \textbf{MTurk question} & \textbf{\% Prefer} & \textbf{\% Prefer} \\
  & \textbf{Our Mode}l& \textbf{Amazon}\\
 \hline
  (A) Image   & 65\%& 35\% \\%& \textbf{0.58} & \textbf{0.67}\\
  (B) Summary  & 30\%&70\% \\ %0.35 &0.48  \\
  (D) Image \& summary pair & 35\% &65\% \\%&0.30 & 0.39\\
  %(E) Emotions/feelings & 37\% & 63\%\\

  \hline
  
\end{tabular}
\caption{Percentage of respondents who prefer our model's automatically generated multimedia summaries over those of Amazon Prime DVD chapter human curated professional summaries.} %Labelers responses to questions 1,2, and 4 related to GoT.
%}  
  \vspace{-1.5em}
\label{tab:GoT-Mturk}
\end{table}

%\begin{table}[h]

%\centering
%\small
%\begin{tabular}{|l|l|l|l|}
%\hline
 %\centering
%\textbf{Method} & \textbf{Reason for selecting summary} & \textbf{Reason for not selecting summary}\\
% MTurk question & Our Model& Amazon key Frame\\
 %\hline
 %Our model Summary & (1) Not opinionated by saying he was showing off & \\
 
%Amazon Prime Summary  & (1) It was more detailed \newline (2) it wasn't the person opinion on how well Beto was doing   &  \\
%\hline

%\end{tabular}
%\caption{Example reasons why Turkers selected a particular summary and why they did not select the other summary: Presidential debate}
%\label{tab:pres-Mturk-qual}
%\end{table}

	\begin{table*}[ht]

\centering
%\small
\begin{tabular}{>{\centering\arraybackslash}m{2.0cm}|>{\centering\arraybackslash}m{6.6cm}|>{\centering\arraybackslash}m{6.5cm}}
\hline
\textbf{Method} & \textbf{Reason for selecting this summary} & \textbf{Reason for \textit{not} selecting this summary} \\
\hline
Our model Summary & %(1) This summary describes more of the content of their conversations \newline (2) This summary best describes the key highlight of the scene/moment \newline (3) This summary was much more entertaining and captures my view of the scene \newline (4) Describes what I was thinking as the scene progressed \newline (5) More entertaining \newline (6)Captures the most important highlight of the scene \newline (7) More entertaining and matches my thoughts of the moment \newline 
(1) I selected this summary as it was the most related to the depiction and while not written in a great way, best described the scene \newline (2) I selected this summary as it seems to better show the feelings and environment of what is being described %\newline (3) I chose this summary as it better depicts the action described \newline (4) I chose this summary as, while entertaining and somewhat silly, provided a greater depiction of what was being described
& (1)  did not choose this summary %the other summary 
because it did not provide enough detail on the scene \newline (2) I did not choose this summary %the other summary 
as it was not descriptive enough\\ %\newline (3) this summary  %the other summary which I did not choose, did not provide any level of detail or mood \newline (4) I did not choose this summary %the other as I did not feel it provided sufficient detail on what was described \\
\hline
Amazon Prime Human Professional Summary  & (1) In this scene the summary best describes the moment and the pending intensity to come \newline (2) This was the most descriptive %\newline (3)Better description of the moment \newline (4) I selected this summary as it seemed to be the most cohesive to the story line that was being described 
& (1)This summary %The second summary 
was too specific to a single moment  \newline %(4) 
(2) This summary %The summary 
not selected portrayed a slightly different aspect of the event. \\

\hline

% \\
%\hline
\end{tabular}
\caption{Qualitative reasons why respondents select a particular summary and why they do not select it: GoT}
  \vspace{-1.5em}
\label{tab:GoT-Mturk-qual}
\end{table*}

%\begin{table}[h]

%\centering
%\small
%\begin{tabular}{|l|l|l|l|}%{lllll}
%\hline
% \centering

 %Amazon vs model & Summary selected& Why other summary was not selected\\\\
 %\hline
  
  %Selected Our model Summary & This summary describes more of the content of their conversations \newline This summary best describes the key highlight of the scene/moment \newline \\
  
  %Selected Amazon Prime Summary  &  In this scene the summary best describes the moment and the pending intensity to come \newline this was the most descriptive\\
  %\hline
%\end{tabular}
%\caption{Qualitative result from Mechanical Turk for GoT}
%\label{tab:GoT-Mturk-qual}
%\end{table}

\noindent
\textbf{Debate:} For each identified scene in the presidential debate, similar to the GoT analysis, we ask 5 respondents - who had watched the presidential debate the following questions: 
\begin{enumerate}[label=(\Alph*)]
\item Given the last NYT reporters live chat in a scene as a summary of the scene and the tweet summary from our multimedia model, select the summary that best \textbf{previews the scene}?
%\paragraph{}
\item State why you selected the summary that you did and what you think about the summary that was not selected (this question was marked as optional)?

%\item  given the last NYT live chat in a sub-event as a summary of the sub-event and the tweet summary from our multimedia model, select the summary that best \textbf{captures the feelings and emotions associated with the scene}.
\end{enumerate}

%we ask the Turkers (which we determined watched the presidential debates) the following questions (1) given the last NYT live-blog post in a sub-event and the tweet summary from our model, select the summary that best previews the sub-event, and (2) state why you selected the summary that you did and what do you think about the summary that was not selected.

For question (A), we select a summary if 3 or more of the respondents selected it. The annotator agreement (using  intraclass correlation) was 0.78. Roughly similar to GoT tweet summaries, the respondents prefer our summaries 33\% of the time over the NYT reporters' live chats summaries. %Also, the respondents thought our summaries captured the feelings and emotions of the scene 54\% of the time compared with the summaries from the NYT reporters.
%The accuracy for the NYT live-blog summaries was 66\% while the accuracy of the summaries from our predicted multimedia summary was 33\%. 
Table \ref{tab:pres-Mturk-qual} shows the qualitative results from the presidential debates. %The quantitative and qualitative results for the presidential debate experiments. 

\begin{table}[h]
\centering
\begin{tabular}{>{\centering\arraybackslash}m{2.5cm}|>{\centering\arraybackslash}m{5.0cm}} \hline
\textbf{Method} & \textbf{Reason for selecting summary}\\ \hline
Our model Summary & (1) Not opinionated by saying he was showing off  \\ \hline
NYT journalist/political analyst live chat & (1) It was more detailed \newline (2) gave more information about the sub-event\\ \hline
\end{tabular}
\caption{Qualitative reasons why respondents select a particular summary.}
  \vspace{-1.5em}
\label{tab:pres-Mturk-qual}
\end{table}

\section{Discussion and Limitations}

%\textbf{State that in the presidential debates data, there were 20 scenes in which the duration between the scene and the previous scene was 1 minute. Our model performed well with scenes that lasted more than 1 minute but did not perform as well with scenes that lasted for 1 minute or less. In the future, we will work on this.}
%\section{Discussion and Limitations}
%In this work, we propose a simple approach to identify and showcase %all 
%the scenes in televised events and we  evaluate our approach on Tweets and videos collected around two events.

In this section we discuss the findings from this work and analyze the limitations and failure cases of our approach to inform future works on this task.

%Regarding scene detection, we observed that our scene detection algorithm identified more scenes in GoT compared to the Presidential debates; this is partly because the GoT scenes tended to be longer (i.e. approximately 4 minutes) while the presidential debates scenes were shorter (i.e. approximately 1.3 minutes).

From the qualitative results, the recurring response as to why the respondents did not select the tweet summary from our model was that some of the tweet summaries were not detailed/descriptive enough or too opinionated when compared to the summaries from Amazon/NYT. However, the fact that tweets are able to capture ``in the moment" feelings and opinions of viewers are also the reason why respondents prefer them over traditional summaries provided by Amazon/NYT.

Figure \ref{examples} shows representative examples of the output from our model.  For GoT, our approach depicts scenes similarly to Amazon Prime. What's more, the descriptions from tweets include not only what happen  e.g., \textit{``The Hound! fighting with Jon!"}, but also provide viewers' general feelings and responses e.g., \textit{``Arya Stark and Hot Pie ... this is so beautiful"}, as well as background information  towards what's happening in the scene e.g., \textit{``Jon was like test me Sansa--I dare you"}, in the scene in which brother and sister, Jon and Sansa, argue over political maneuvers. Similarly, for the debate, our multimedia summaries depict descriptions of the scenes and include feelings/opinions of the viewers toward the scenes e.g., \textit{"I give much credit to @juliancastro for being the first to talk about transgender issues on the debate stage. \#dncdebate"}.

%I feel the other summary is too opinion based

%	\begin{table*}[ht]

%\centering
%\small
%\begin{tabular}{>{\centering\arraybackslash}m{2.0cm}|>{\centering\arraybackslash}m{6.6cm}|>{\centering\arraybackslash}m{6.5cm}}
%\hline
%\textbf{Method} & \textbf{Reason for selecting summary} & \textbf{Reason for not selecting summary} \\
%\hline
%Our model Summary & (1) Not opinionated by saying he was showing off & \\

%hline
%Amazon Prime Summary  & (1) It was more detailed \newline(2) it wasn't the person opinion on how well Beto was doing   &  \\
%\hline

% \\
%\hline
%\end{tabular}
%\caption{Example reasons why Turkers selected a particular summary and why they did not select the other summary: Presidential debate}
%\label{tab:pres-Mturk-qual}
%\end{table*}

%multimedia summary (i.e. image and tweet) predicted by our model and the Amazon Prime key frame and short description, we asked Turkers to (i) select the either the image from our predicted multimedia summary or the key frame We qualitatively compared our model's multimedia summary results against the scene description and pictures that we extracted from Amazon Prime for GoT and the . \textbf{Andy: write about mturk task for presidential debate and GoT}

By selecting video frames that contain characters mentioned in these tweets, our method is able to select pictures that are relevant to the tweet summaries and are comparable to Amazon Prime frames or more informative e.g., the frame where the character, Euron is walking into king's landing with captives. %or the frame where Daenerys is going to punish the defeated soldiers. 
Taken together, we observe for both the presidential debate and GoT that more than one-third of annotators prefer our \textit{multimedia} summaries as they reflect the opinions and reactions of Twitter users despite the summaries coming from tweets that are likely not as coherent or as well written as the gold standard, human expert and professional summaries from NYT and Amazon Prime.

A potential use case of our model is to summarize and capture diverse user opinions (from social media posts) about events in real-time.\\ 

\noindent
\textbf{Limitations and Error Analysis:}
%Our approach has some limitations and some failure cases; below we discuss them.
Firstly, for our scene identification step, the most common scenes that this approach missed were sequential scenes that involved the same character; this happened 8 times in GoT. This is partly because our approach assumes that the character that determines a new scene will be different from the characters from the previous scene, however, in these events, some adjacent scenes involve the same characters. 

Another failure case comes from when our face recognition model predicts the name of the face wrongly, %(Figure \ref{wrong}), 
which results in a wrong frame being selected. This happens in 9 of the 76 ($\sim$12\%) identified scenes in our GoT dataset. Another error comes from the wrong name being identified in the %correct 
tweet of the scene: %(Figure \ref{wrong2}: 
we mistake the Twitter user name as containing a mention of a character, which results in the wrong frame being selected. This happens in 1 of the 76 identified scenes ($\sim$1\%) in GoT dataset. 

%\begin{figure}[h]
%\centering
%\includegraphics[scale=0.36]{wrong.png}
%		\caption{Wrongly predicted names for  faces results in incorrect frame selection (wrong faces marked)}
%		\label{wrong}
%		\vspace{-1.5em}
%\end{figure}

%\begin{figure}[h]
%\centering
%\includegraphics[scale=0.35]{wrong2.png}
%		\caption{The name ``dany" is incorrectly identified from the tweet, which results in the wrong frame containing dany's face being selected. The correct frame, showing the undead scrambling in mass towards the group of men -- indeed like ``the black friday sale" -- is shown on the right}
%		\label{wrong2}
%		\vspace{-1.5em}
%\end{figure}

We also observe that if characters appear together in all episodes, our approach can wrongly predict their labels. For example, \textit{Lyanna} and \textit{Robert} is wrongly predicted as each other as they only ever appear together in all GoT season 7 episodes. %(Figure \ref{failure}). 
We believe exposing our model to more episodes can help in this case. 

Lastly, in this work, we focused on popular events which generate a lot of Twitter data from viewers, hence, the findings from this work may not apply to events with fewer associated Twitter posts.

%, hence, the findings from this work may not apply to less popular events which may not generate as much social media data as the popular events.

%\textbf{state that a limitation is that we worked with popular events and in the future we hope to work with less popular events that generate less tweets.}
%	\begin{figure}[h]
		
%	\includegraphics[width=0.48\textwidth]{failures.png}
%		\caption{The top-10 faces predicted by our model to be \textit{Lyanna}'s and \textit{Robett}'s show how our model can wrongly label characters who \textit{always} appear together throughout the show.}
%		\label{failure}
%		\vspace{1.5em}
%	\end{figure}

%\section{Ethics and Privacy}

%In this work, although we did not use the Twitter user names of the authors of the posts for our experiments, for privacy, we de-identified the Twitter user names. %of the authors of the Twitter posts.

%\section{Discussion}
%\textbf{Andy: write about selecting more that one tweet as a summary of the sub-event}

%\noindent
%\begin{center}
%\textbf{Conclusion and Future Work}
%\end{center}
%\noindent

\section{Conclusion and Future Work}

%\section{Conclusion and Future Work}
% talk about other potential things we can use with real live comments and video (other sites like Viki made this possible) and using comments from multiple languages referring to the same scenes could be interesting for other purposes like machine translation 

In conclusion, we propose a simple approach to identify and portray the scenes in televised events. 
We evaluate our approach on tweets collected around 7 episodes of a TV-show, GoT and night 1 of the first 
Democratic party presidential debates. %We conduct quantitative and qualitative evaluation, comparing our summaries with real life human expert and professional summaries of these events, and show that our approach  works well for detecting and depicting sequential sub-events with images and descriptions from videos and tweets, particularly for capturing \textit{in the moment} feelings and reactions of viewers. As our summary can effectively capture the opinions and reactions of Twitter users to different sub-events of an event, it can be beneficial for gauging how people are reacting to the event while also providing live reports of the event. 
To the best of our knowledge, our work is the first to combine knowledge from multiple modalities \textit{and} platforms of different nature i.e., \textit{narration} and \textit{reaction} about the event in Twitter and \textit{dialogues} in the televised video of the event, to obtain coherent and sequential portrayal of scenes in televised events. 

%In this work, we 
This work uses English Twitter data, however, there are events such as %popular sports events like  
the Olympics that attract users -- from various countries and that speak different languages, some of whom  publish Twitter data in their respective languages about the event in real-time. In the future, we envision the use of our method for  %collecting social media data, in different languages, related to an event and 
creating real-time multilingual and multimedia summaries of events around the world using videos and social media data. Such multimedia summaries can also be useful for downstream applications such as sentiment analysis of users' reactions toward events and machine translation (MT) of UGCs. %User Generated Contents (UGCs). %, which is an important yet challenging task due to the informal and noisy nature of UGCs data and the lack of UGCs training data for MT \cite{michel2018mtnt}. 
As social media users may talk about the same event similarly in different languages, we can use the multilingual and multimedia summaries of their responses as comparable data, with shared images as additional information, for training future MT models for UGCs. 

\bibliographystyle{aaai}
\bibliography{ref}

\end{document}